\documentclass{article}

\usepackage{times}
\usepackage{graphicx} 
\usepackage{subfigure} 

\usepackage{natbib}

\usepackage{algorithm}
\usepackage{algorithmic}

\usepackage{xspace}
\usepackage{amsmath,amssymb,amsfonts,comment}
\usepackage{graphics,graphicx}
\usepackage{dsfont}
\usepackage{afterpage}

\usepackage[accepted]{icml2016}

\icmltitlerunning{Dynamic Memory Networks for Visual and Textual Question Answering}

\begin{document} 

\twocolumn[

\icmltitle{Dynamic Memory Networks for Visual and Textual Question Answering}

\icmlauthor{Caiming Xiong*, Stephen Merity*, Richard Socher}{\{cmxiong,smerity,richard\}\@metamind.io}
\icmladdress{MetaMind, Palo Alto, CA USA \hspace{5.5cm}*indicates equal contribution.}

\icmlkeywords{deep learning, attention}

\vskip 0.3in
]

\newcommand{\babi}{bAbI\xspace}

\begin{abstract} 
Neural network architectures with memory and attention mechanisms exhibit certain reasoning capabilities required for question answering.
One such architecture, the dynamic memory network (DMN), obtained high accuracy on a variety of language tasks.
However, it was not shown whether the architecture achieves strong results for question answering when supporting facts are not marked during training or whether it could be applied to other modalities such as images.
Based on an analysis of the DMN, we propose several improvements to its memory and input modules. Together with these changes we introduce a novel input module for images in order to be able to answer visual questions.
Our new DMN+ model improves the state of the art on both the Visual Question Answering dataset and the \babi-10k text question-answering dataset without supporting fact supervision.
\end{abstract} 

\section{Introduction}
Neural network based methods have made tremendous progress in image and text classification \cite{Krizhevsky2012,Socher2013EMNLP}.
However, only recently has progress been made on more complex tasks that require logical reasoning. This success is based in part on the addition of memory and attention components to complex neural networks. For instance, memory networks \cite{Weston2015} are able to reason over several facts written in natural language or (subject, relation, object) triplets. Attention mechanisms have been successful components in both machine translation \cite{Bahdanau2015,Luong2015} and image captioning models \cite{Xu2015}.

The dynamic memory network \cite{Kumar2015} (DMN) is one example of a neural network model that has both a memory component and an attention mechanism.
The DMN yields state of the art results on question answering with supporting facts marked during training, sentiment analysis, and part-of-speech tagging.

\begin{figure}[t]
\centering
\includegraphics[scale=0.4]{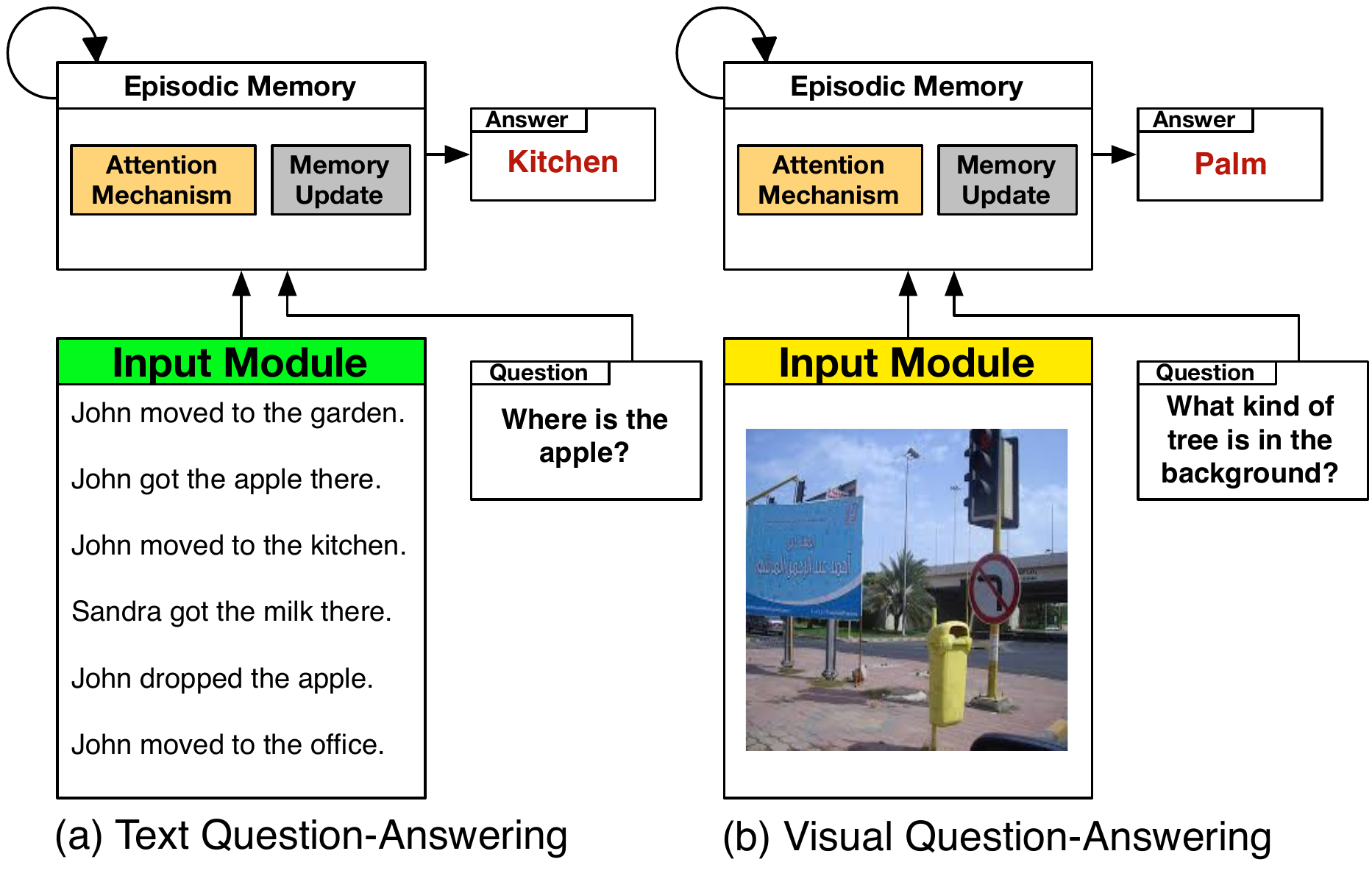}
\vspace{-0.3cm}
\caption{Question Answering over text and images using a Dynamic Memory Network.}
\vspace{-0.6cm}
\label{fig:fig1}
\end{figure}

We analyze the DMN components, specifically the input module and memory module, to improve question answering.
We propose a new input module which uses a two level encoder with a sentence reader and input fusion layer to allow for information flow between sentences. 
For the memory, we propose a modification to gated recurrent units (GRU) \cite{Chung2014}. The new GRU formulation incorporates attention gates  that are computed using global knowledge over the facts.
Unlike before, the new DMN+ model does not require that supporting facts (i.e. the facts that are relevant for answering a particular question) are labeled during training. The model learns to select the important facts from a larger set.

In addition, we introduce a new input module to represent images. This module is compatible with the rest of the DMN architecture and its output is fed into the memory module. We show that the changes in the memory module that improved textual question answering also improve visual question answering. Both tasks are illustrated in Fig.~\ref{fig:fig1}. 

\section{Dynamic Memory Networks}
We begin by outlining the DMN for question answering and the modules as presented in \citet{Kumar2015}. 

The DMN is a general architecture for question answering (QA). It is composed of modules that allow different aspects such as input representations or memory components to be analyzed and improved independently. 
The modules, depicted in Fig.~\ref{fig:fig1}, are as follows:

\textbf{Input Module}:
This module processes the input data about which a question is being asked into a set of vectors termed facts, represented as $F=[f_1,\hdots,f_N]$, where $N$ is the total number of facts.
These vectors are ordered, resulting in additional information that can be used by later components.
For text QA in \citet{Kumar2015}, the module consists of a GRU over the input words.

As the GRU is used in many components of the DMN, it is useful to provide the full definition.
For each time step $i$ with input $x_i$ and previous hidden state $h_{i-1}$, we compute the updated hidden state $h_i = GRU(x_i,h_{i-1})$ by
\begin{eqnarray}
u_i &=& \sigma\left(W^{(u)}x_{i} + U^{(u)} h_{i-1}  + b^{(u)} \right)\label{eq:gru-update}\\
r_i &=& \sigma\left(W^{(r)}x_{i} + U^{(r)} h_{i-1} + b^{(r)} \right)\\
\tilde{h}_i &=&  \tanh\left(Wx_{i} + r_i \circ U h_{i-1}  + b^{(h)}\right)\\
h_i &=&  u_i\circ \tilde{h}_i + (1-u_i) \circ h_{i-1}\label{eq:gru-hidden}
\end{eqnarray}
where $\sigma$ is the sigmoid activation function, $\circ$ is an element-wise product, $W^{(z)}, W^{(r)}, W \in \mathbb{R}^{n_H \times n_I}$, $U^{(z)}, U^{(r)}, U \in \mathbb{R}^{n_H \times n_H}$, $n_H$ is the hidden size, and $n_I$ is the input size.


\textbf{Question Module}:
This module computes a vector representation $q$ of the question, where $q \in \mathbb{R}^{n_H}$ is the final hidden state of a GRU over the words in the question.

\textbf{Episodic Memory Module}:
Episode memory aims to retrieve the information required to answer the question $q$ from the input facts.
To improve our understanding of both the question and input, especially if questions require transitive reasoning, the episode memory module may pass over the input multiple times, updating episode memory after each pass.
We refer to the episode memory on the $t^{th}$ pass over the inputs as $m^t$, where $m^t \in \mathbb{R}^{n_H}$, the initial memory vector is set to the question vector: $m^0 = q$.

The episodic memory module consists of two separate components: the attention mechanism and the memory update mechanism.
The attention mechanism is responsible for producing a contextual vector $c^t$, where $c^t \in \mathbb{R}^{n_H}$ is a summary of relevant input for pass $t$, with relevance inferred by the question $q$ and previous episode memory $m^{t-1}$.
The memory update mechanism is responsible for generating the episode memory $m^t$ based upon the contextual vector $c^t$ and previous episode memory $m^{t-1}$.
By the final pass $T$, the episodic memory $m^T$ should contain all the information required to answer the question $q$.

\textbf{Answer Module}:
The answer module receives both $q$ and $m^T$ to generate the model's predicted answer.
For simple answers, such as a single word, a linear layer with softmax activation may be used.
For tasks requiring a sequence output, an RNN may be used to decode $a = [q ; m^T]$, the concatenation of vectors $q$ and $m^T$,  to an ordered set of tokens.
The cross entropy error on the answers is used for training and backpropagated through the entire network.

\section{Improved Dynamic Memory Networks: DMN+}
We propose and compare several modeling choices for two crucial components: input representation, attention mechanism and memory update. The final DMN+ model obtains the highest accuracy on the \babi-10k dataset without supporting facts and the VQA dataset \cite{Antol2015}. Several design choices are motivated by intuition and accuracy improvements on that dataset. 

\subsection{Input Module for Text QA} \label{inputTQA}
In the DMN specified in \citet{Kumar2015}, a single GRU is used to process all the words in the story, extracting sentence representations by storing the hidden states produced at the end of sentence markers.
The GRU also provides a temporal component by allowing a sentence to know the content of the sentences that came before them.
Whilst this input module worked well for \babi-1k with supporting facts, as reported in \citet{Kumar2015}, it did not perform well on \babi-10k without supporting facts (Sec. \ref{sec:model-analysis}).

We speculate that there are two main reasons for this performance disparity, all exacerbated by the removal of supporting facts.
First, the GRU only allows sentences to have context from sentences before them, but not after them.
This prevents information propagation from future sentences.
Second, the supporting sentences may be too far away from each other on a word level to allow for these distant sentences to interact through the word level GRU.

\textbf{Input Fusion Layer} \label{sec:fusion}

For the DMN+, we propose replacing this single GRU with two different components.
The first component is a sentence reader, responsible only for encoding the words into a sentence embedding.
The second component is the input fusion layer, allowing for interactions between sentences.
This resembles the hierarchical neural auto-encoder architecture of \citet{Li2015} and allows content interaction between sentences.
We adopt the bi-directional GRU for this input fusion layer because it allows information from both past and future sentences to be used.
As gradients do not need to propagate through the words between sentences, the fusion layer also allows for distant supporting sentences to have a more direct interaction.

\begin{figure}
\centering
\includegraphics[width=0.48\textwidth]{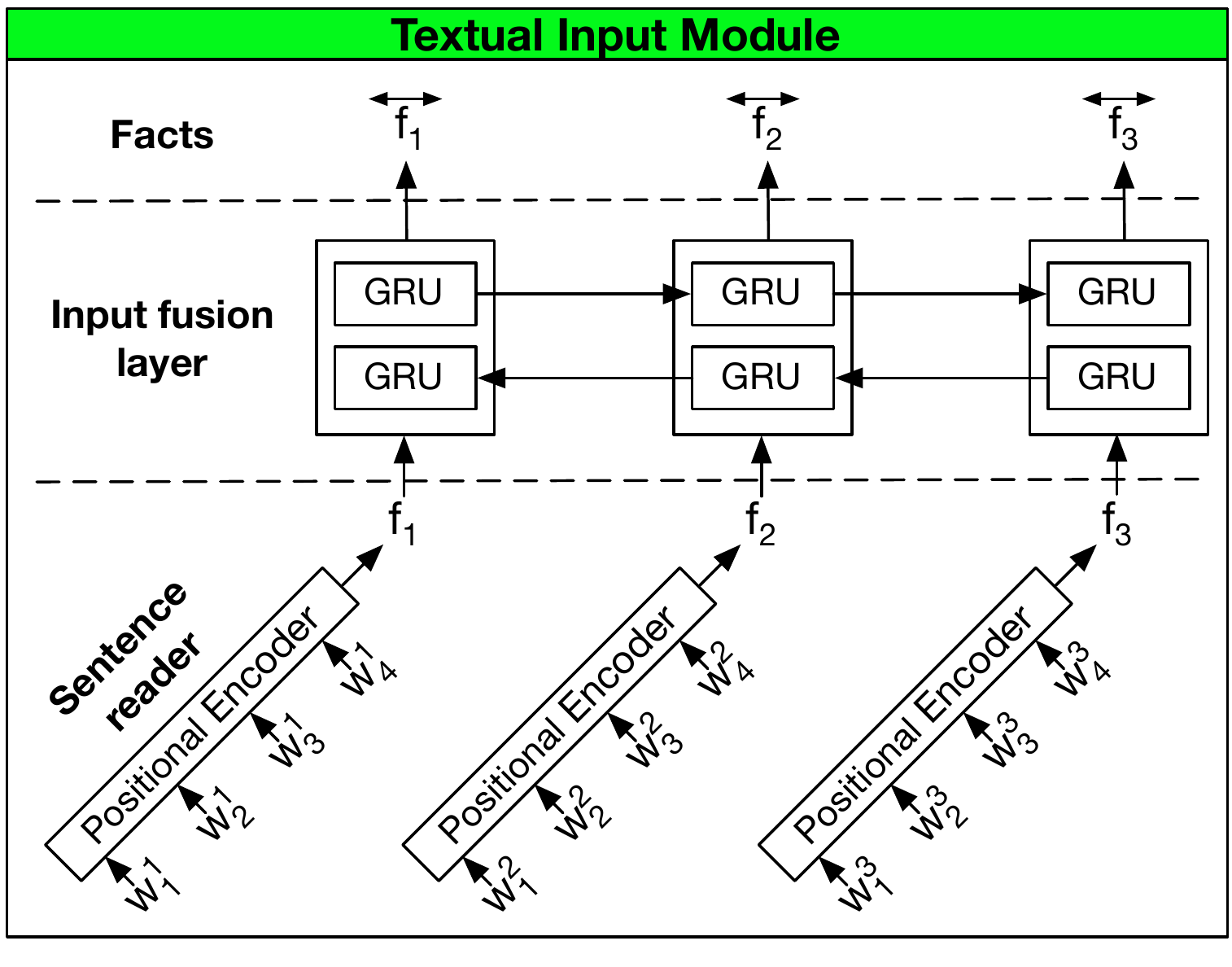}
\vspace{-0.3cm}
\caption{The input module with a ``fusion layer'', where the sentence reader encodes the sentence and the bi-directional GRU allows information to flow between sentences.}
\vspace{-0.3cm}
\label{fig:fusion}
\end{figure}

Fig.~\ref{fig:fusion} shows an illustration of an input module, where a positional encoder is used for the sentence reader and a bi-directional GRU is adopted for the input fusion layer.
Each sentence encoding $f_i$ is the output of an encoding scheme taking the word tokens $[w^i_1, \hdots, w^i_{M_i}]$, where $M_i$ is the length of the sentence.

The sentence reader could be based on any variety of encoding schemes. We selected positional encoding described in \citet{Sukhbaatar2015} to allow for a comparison to their work.
GRUs and LSTMs were also considered but required more computational resources and were prone to overfitting if auxiliary tasks, such as reconstructing the original sentence, were not used.

For the positional encoding scheme, the sentence representation is produced by $f_i = \sum^{j=1}_M l_j \circ w^i_j$, where $\circ$ is element-wise multiplication and $l_j$ is a column vector with structure $l_{jd} = (1 - j / M) - (d / D) (1 - 2j / M)$, where $d$ is the embedding index and $D$ is the dimension of the embedding.


The input fusion layer takes these input facts and enables an information exchange between them by applying a bi-directional GRU.
\begin{eqnarray}
\overrightarrow{f_i} = GRU_{fwd}(f_i, \overrightarrow{f_{i-1}}) \\
\overleftarrow{f_{i}} = GRU_{bwd}(f_{i}, \overleftarrow{f_{i+1}}) \\
\overleftrightarrow{f_i} = \overleftarrow{f_i} + \overrightarrow{f_i}
\end{eqnarray}
where $f_i$ is the input fact at timestep $i$, $ \overrightarrow{f_i}$ is the hidden state of the forward GRU at timestep $i$, and $\overleftarrow{f_i}$ is the hidden state of the backward GRU at timestep $i$.
This allows contextual information from both future and past facts to impact $\overleftrightarrow{f_i}$.

We explored a variety of encoding schemes for the sentence reader, including GRUs, LSTMs, and the positional encoding scheme described in \citet{Sukhbaatar2015}.
For simplicity and speed, we selected the positional encoding scheme.
GRUs and LSTMs were also considered but required more computational resources and were prone to overfitting if auxiliary tasks, such as reconstructing the original sentence, were not used.

\subsection{Input Module for VQA} \label{inputVQA}
\begin{figure}
   \centering
   	 	\includegraphics[width=0.5\textwidth]{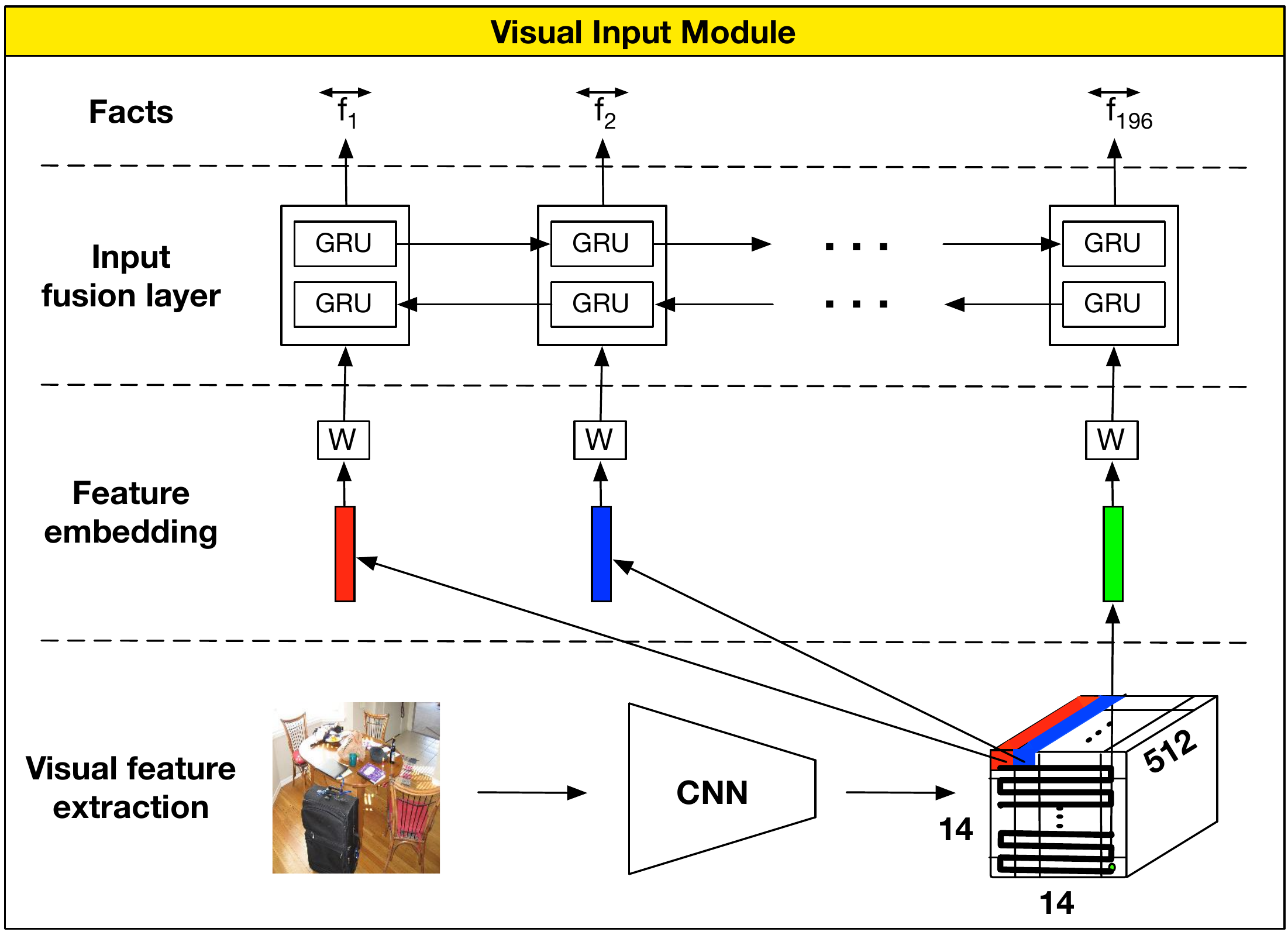}
   	 	\vspace{-0.4cm}
   \caption{VQA input module to represent images for the DMN.}\label{fig:result}
   \label{fig:vqa}
\end{figure}

To apply the DMN to visual question answering, we introduce a new input module for images. The module splits an image into small local regions and considers each region equivalent to a sentence in the input module for text.
The input module for VQA is composed of three parts, illustrated in Fig.~\ref{fig:vqa}: local region feature extraction, visual feature embedding, and the input fusion layer introduced in Sec.~\ref{sec:fusion}.

\textbf{Local region feature extraction:} 
To extract features from the image, we use a convolutional neural network \cite{Krizhevsky2012} based upon the VGG-19 model \cite{simonyan2014very}.
We first rescale the input image to $448 \times 448$ and take the output from the last pooling layer which has dimensionality $d = 512 \times 14 \times 14$.
The pooling layer divides the image into a grid of $14 \times 14$, resulting in 196 local regional vectors of $d = 512$.

\textbf{Visual feature embedding:} 
As the VQA task involves both image features and text features, we add a linear layer with tanh activation to project the 
local regional vectors to the textual feature space used by the question vector $q$.

\textbf{Input fusion layer:} 
The local regional vectors extracted from above do not yet have global information available to them.
Without global information, their representational power is quite limited, with simple issues like object scaling or locational variance causing accuracy problems.

To solve this, we add an input fusion layer similar to that of the textual input module described in Sec.~\ref{inputTQA}.
First, to produce the input facts $F$, we traverse the image in a snake like fashion, as seen in Figure \ref{fig:vqa}.
We then apply a bi-directional GRU over these input facts $F$ to produce the globally aware input facts $\overleftrightarrow{F}$.
The bi-directional GRU allows for information propagation from neighboring image patches, capturing spatial information.

\subsection{The Episodic Memory Module}
\begin{figure}
\centering
\includegraphics[height=5cm]{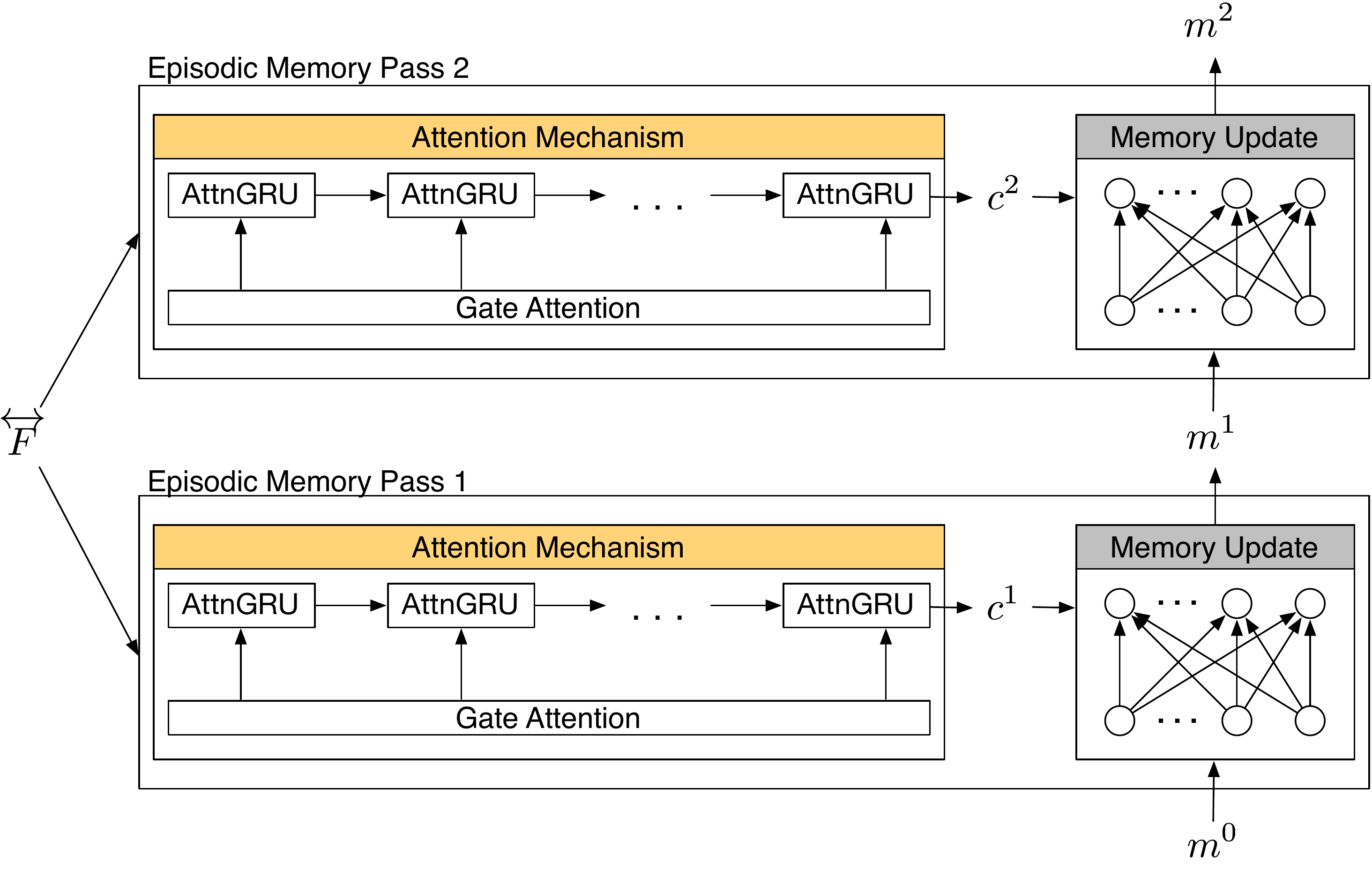}
\vspace{-0.3cm}
\caption{The episodic memory module of the DMN+ when using two passes. The $\overleftrightarrow{F}$ is the output of the input module.
\label{fig:episodicModule}
}
\end{figure}

The episodic memory module, as depicted in Fig.~\ref{fig:episodicModule}, retrieves information from the input facts $\overleftrightarrow{F} = [\overleftrightarrow{f_1}, \hdots, \overleftrightarrow{f_N}]$ provided to it by focusing attention on a subset of these facts.
We implement this attention by associating a single scalar value, the attention gate $g^t_i$, with each fact $\overleftrightarrow{f}_i$ during pass $t$.
This is computed by allowing interactions between the fact and both the question representation and the episode memory state.
\setlength\arraycolsep{0.4pt}
\begin{eqnarray}
z^t_i &=& [\overleftrightarrow{f_i} \circ q; \overleftrightarrow{f_i} \circ m^{t-1}; \lvert \overleftrightarrow{f_i} - q \rvert; \lvert \overleftrightarrow{f_i} - m^{t-1} \rvert]\label{eq:z} \\
Z^t_i &=& W^{(2)} \tanh\left(W^{(1)}z^t_i + b^{(1)} \right)+ b^{(2)} \\
g^t_i &=& \frac{\exp(Z^t_i)}{\sum_{k=1}^{M_i} \exp(Z^t_k)} \label{eq:attn-gate}
\end{eqnarray}
where $\overleftrightarrow{f_i}$ is the $i^{th}$ fact, $m^{t-1}$ is the previous episode memory, $q$ is the original question, $\circ$ is the element-wise product, $|\cdot|$ is the element-wise absolute value, and $;$ represents concatenation of the vectors.

The DMN implemented in \citet{Kumar2015} involved a more complex set of interactions within $z$, containing the additional terms $[f; m^{t-1}; q; f^T W^{(b)} q; f^T W^{(b)} m^{t-1}]$.
After an initial analysis, we found these additional terms were not required.

\textbf{Attention Mechanism}

Once we have the attention gate $g^t_i$ we use an attention mechanism to extract a contextual vector $c^t$ based upon the current focus.
We focus on two types of attention: soft attention and a new attention based GRU. The latter improves performance and is hence the final modeling choice for the DMN+.

\textbf{Soft attention:}
Soft attention produces a contextual vector $c^t$ through a weighted summation of the sorted list of vectors $\overleftrightarrow{F}$ and corresponding attention gates $g_i^t$: $c^t = \sum_{i=1}^N g^t_i \overleftrightarrow{f}_i$
This method has two advantages.
First, it is easy to compute.
Second, if the softmax activation is spiky it can approximate a hard attention function by selecting only a single fact for the contextual vector whilst still being differentiable.
However the main disadvantage to soft attention is that the summation process loses both positional and ordering information.
Whilst multiple attention passes can retrieve some of this information, this is inefficient.

\textbf{Attention based GRU:} \label{sec:attngru}
For more complex queries, we would like for the attention mechanism to be sensitive to both the position and ordering of the input facts $\overleftrightarrow{F}$.
An RNN would be advantageous in this situation except they cannot make use of the attention gate from Equation \ref{eq:attn-gate}.

We propose a modification to the GRU architecture by embedding information from the attention mechanism.
The update gate $u_i$ in Equation \ref{eq:gru-update} decides how much of each dimension of the hidden state to retain and how much should be updated with the transformed input $x_i$ from the current timestep.
As $u_i$ is computed using only the current input and the hidden state from previous timesteps, it lacks any knowledge from the question or previous episode memory.

\begin{figure}
\centering
\includegraphics[height=2.5cm]{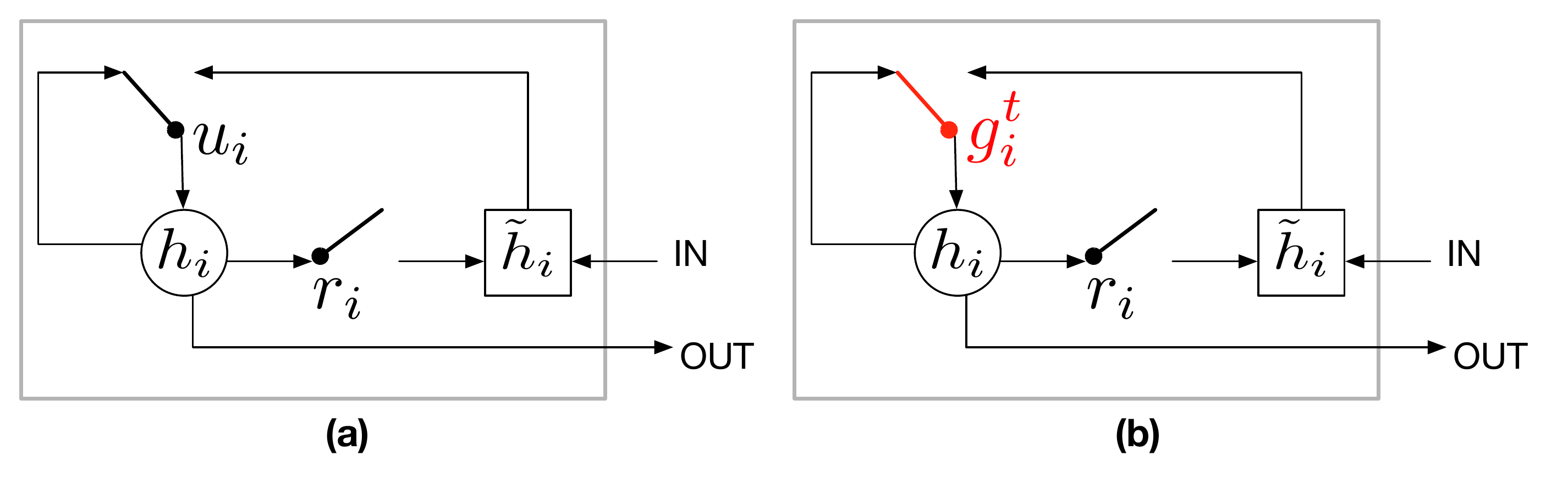}
\vspace{-0.3cm}
\caption{(a) The traditional GRU model, and (b) the proposed attention-based GRU model}
\vspace{-0.3cm}
\label{fig:attention}
\end{figure}

By replacing the update gate $u_i$ in the GRU (Equation \ref{eq:gru-update}) with the output of the attention gate $g^t_i$ (Equation \ref{eq:attn-gate}) in Equation \ref{eq:gru-hidden}, the GRU can now use the attention gate for updating its internal state.
This change is depicted in Fig~\ref{fig:attention}.
\begin{eqnarray}
h_i &=&  g^t_i \circ \tilde{h}_i + (1-g^t_i) \circ h_{i-1}
\end{eqnarray}
An important consideration is that $g^t_i$ is a scalar, generated using a softmax activation, as opposed to the vector $u_i \in \mathbb{R}^{n_H}$, generated using a sigmoid activation.
This allows us to easily visualize how the attention gates activate over the input, later shown for visual QA in Fig.~\ref{fig:qualitative}.
Though not explored, replacing the softmax activation in Equation \ref{eq:attn-gate} with a sigmoid activation would result in $g^t_i \in \mathbb{R}^{n_H}$.
To produce the contextual vector $c^t$ used for updating the episodic memory state $m^t$, we use the final hidden state of the attention based GRU.

\textbf{Episode Memory Updates}

After each pass through the attention mechanism, we wish to update the episode memory $m^{t-1}$ with the newly constructed contextual vector $c^t$, producing $m^t$.
In the DMN, a GRU with the initial hidden state set to the question vector $q$ is used for this purpose.
The episodic memory for pass $t$ is computed by
\begin{equation}
m^t = GRU(c^t, m^{t-1})
\end{equation}

The work of \citet{Sukhbaatar2015} suggests that using different weights for each pass through the episodic memory may be advantageous.
When the model contains only one set of weights for all episodic passes over the input, it is referred to as a \textbf{tied model}, as in the ``Mem Weights'' row in Table~\ref{table:babi-compare}.


Following the memory update component used in \citet{Sukhbaatar2015} and \citet{Peng2015} we experiment with using a ReLU layer for the memory update, calculating the new episode memory state by
\begin{equation}
m^t = ReLU\left(W^t [m^{t-1} ; c^t ; q] + b\right)
\end{equation}
where $;$ is the concatenation operator, $W^t \in \mathbb{R}^{n_H \times n_H}$, $b \in \mathbb{R}^{n_H}$, and $n_H$ is the hidden size.
The untying of weights and using this ReLU formulation for the memory update improves accuracy by another 0.5\% as shown in Table~\ref{table:babi-compare} in the last column.
The final output of the memory network is passed to the answer module as in the original DMN.

\section{Related Work}
The DMN is related to two major lines of recent work: memory and attention mechanisms.
We work on both visual and textual question answering which have, until now, been developed in separate communities.

\textbf{Neural Memory Models}
The earliest recent work with a memory component that is applied to language processing is that of memory networks \cite{Weston2015} which adds a memory component for question answering over simple facts.
They are similar to DMNs in that they also have input, scoring, attention and response mechanisms.
However, unlike the DMN their input module computes sentence representations independently and hence cannot easily be used for other tasks such as sequence labeling.
Like the original DMN, this memory network requires that supporting facts are labeled during QA training.
End-to-end memory networks \cite{Sukhbaatar2015} do not have this limitation.
In contrast to previous memory models with a variety of different functions for memory attention retrieval and representations, DMNs \cite{Kumar2015} have shown that neural sequence models can be used for input representation, attention and response mechanisms. 
Sequence models naturally capture position and temporality of both the inputs and transitive reasoning steps.

\textbf{Neural Attention Mechanisms} Attention mechanisms allow neural network models to use a question to selectively pay attention to specific inputs.
They can benefit image classification \cite{Stollenga2014}, generating captions for images \cite{Xu2015}, among others mentioned below, and machine translation \cite{Cho2014b,Bahdanau2015,Luong2015}.
Other recent neural architectures with memory or attention which have
proposed include neural Turing machines \cite{Graves2014}, neural GPUs \cite{Kaiser2015} and stack-augmented RNNs \cite{Joulin2015}.

\textbf{Question Answering in NLP} 
Question answering involving natural language can be solved in a variety of ways to which we cannot all do justice. If the potential input is a large text corpus, QA becomes a combination of information retrieval and extraction \cite{Yates2007}.
Neural approaches can include reasoning over knowledge bases, \cite{Bordes2012,Socher2013:NTN} or directly via sentences for trivia competitions \cite{Iyyer2014}. 

\textbf{Visual Question Answering (VQA)}
In comparison to QA in NLP, VQA is still a relatively young task that is feasible only now that objects can be identified with high accuracy.
The first large scale database with unconstrained questions about images was introduced by \citet{Antol2015}.
While VQA datasets existed before they did not include open-ended, free-form questions about general images \cite{Geman2014}.
Others are were too small to be viable for a deep learning approach \cite{Malinowski2014}. 
The only VQA model which also has an attention component is the stacked attention network \cite{yang2015stacked}.
Their work also uses CNN based features.
However, unlike our input fusion layer, they use a single layer neural network to map the features of each patch to the dimensionality of the question vector.
Hence, the model cannot easily incorporate adjacency of local information in its hidden state.
A model that also uses neural modules, albeit logically inspired ones, is that by \citet{andreas2016learning} who evaluate on knowledgebase reasoning and visual question answering. We compare directly to their method on the latter task and dataset.

Related to visual question answering is the task of describing images with sentences \cite{Kulkarni11}.
\citet{Socher2014TACL} used deep learning methods to map images and sentences into the same space in order to describe images with sentences and to find images that best visualize a sentence.
This was the first work to map both modalities into a joint space with deep learning methods, but it could only select an existing sentence to describe an image. Shortly thereafter, recurrent neural networks were used to generate often novel sentences based on images \cite{Karpathy2015,Chen2014,Fang2015,Xu2015}.

\section{Datasets}

To analyze our proposed model changes and compare our performance with other architectures, we use three datasets.

\subsection{\babi-10k}

For evaluating the DMN on textual question answering, we use \babi-10k English \cite{Weston2015ToyTasks}, a synthetic dataset which features 20 different tasks.
Each example is composed of a set of facts, a question, the answer, and the supporting facts that lead to the answer.
The dataset comes in two sizes, referring to the number of training examples each task has: \babi-1k and \babi-10k.
The experiments in \citet{Sukhbaatar2015} found that their lowest error rates on the smaller \babi-1k dataset were on average three times higher than on \babi-10k.

\subsection{DAQUAR-ALL visual dataset}

The DAtaset for QUestion Answering on Real-world images (DAQUAR) \cite{Malinowski2014} consists of 795 training images and 654 test images.
Based upon these images, 6,795 training questions and 5,673 test questions were generated.
Following the previously defined experimental method, we exclude multiple word answers \cite{malinowski2015ask,ma2015learning}.
The resulting dataset covers 90\% of the original data.
The evaluation method uses classification accuracy over the single words.
We use this as a development dataset for model analysis (Sec. \ref{sec:model-analysis}).

\subsection{Visual Question Answering}
The Visual Question Answering (VQA) dataset was constructed using the Microsoft COCO dataset \cite{lin2014microsoft} which contained 123,287 training/validation images and 81,434 test images.
Each image has several related questions with each question answered by multiple people.
This dataset contains 248,349 training questions, 121,512 validation questions, and 244,302 for testing.
The testing data was split into test-development, test-standard and test-challenge in \citet{Antol2015}.

Evaluation on both test-standard and test-challenge are implemented via a submission system.
test-standard may only be evaluated 5 times and test-challenge is only evaluated at the end of the competition.
To the best of our knowledge, VQA is the largest and most complex image dataset for the visual question answering task.

\section{Experiments}
\subsection{Model Analysis} \label{sec:model-analysis}

To understand the impact of the proposed module changes, we analyze the performance of a variety of DMN models on textual and visual question answering datasets.

The original DMN (ODMN) is the architecture presented in \citet{Kumar2015} without any modifications.
DMN2 only replaces the input module with the input fusion layer (Sec.~\ref{sec:fusion}).
DMN3, based upon DMN2, replaces the soft attention mechanism with the attention based GRU proposed in Sec.~\ref{sec:attngru}.
Finally, DMN+, based upon DMN3, is an untied model, using a unique set of weights for each pass and a linear layer with a ReLU activation to compute the memory update.
We report the performance of the model variations in Table \ref{table:babi-compare}.

A large improvement to accuracy on both the \babi-10k textual and DAQUAR visual datasets results from updating the input module, seen when comparing ODMN to DMN2.
On both datasets, the input fusion layer improves interaction between distant facts.
In the visual dataset, this improvement is purely from providing contextual information from neighboring image patches, allowing it to handle objects of varying scale or questions with a locality aspect.
For the textual dataset, the improved interaction between sentences likely helps the path finding required for logical reasoning when multiple transitive steps are required.

The addition of the attention GRU in DMN3 helps answer questions where complex positional or ordering information may be required.
This change impacts the textual dataset the most as few questions in the visual dataset are likely to require this form of logical reasoning.
Finally, the untied model in the DMN+ overfits on some tasks compared to DMN3, but on average the error rate decreases.

From these experimental results, we find that the combination of all the proposed model changes results, culminating in DMN+, achieves the highest performance across both the visual and textual datasets.

\begin{table}
\centering
{\small
\begin{tabular}{lrrrr}
Model & ODMN & DMN2 & DMN3 & DMN+ \\
\hline
Input module & GRU & Fusion & Fusion & Fusion \\
Attention & $\sum g_i f_i$ & $\sum g_i f_i$ & AttnGRU & AttnGRU \\
Mem update & GRU & GRU & GRU & ReLU \\
Mem Weights & Tied & Tied & Tied & Untied \\
\hline
\multicolumn{5}{c}{\babi English 10k dataset}\\
\hline
QA2 & 36.0 & 0.1 & 0.7 & 0.3\\
QA3 & 42.2 & 19.0 & 9.2 & 1.1\\
QA5 & 0.1 & 0.5 & 0.8 & 0.5\\
QA6 & 35.7 & 0.0 & 0.6 & 0.0\\
QA7 & 8.0 & 2.5 & 1.6 & 2.4\\
QA8 & 1.6 & 0.1 & 0.2 & 0.0\\
QA9 & 3.3 & 0.0 & 0.0 & 0.0\\
QA10 & 0.6 & 0.0 & 0.2 & 0.0\\
QA14 & 3.6 & 0.7 & 0.0 & 0.2\\
QA16 & 55.1 & 45.7 & 47.9 & 45.3\\
QA17 & 39.6 & 5.9 & 5.0 & 4.2\\
QA18 & 9.3 & 3.8 & 0.1 & 2.1\\
QA20 & 1.9 & 0.0 & 0.0 & 0.0\\
\hline
Mean error & 11.8 & 3.9 & 3.3 & 2.8\\
\hline
\multicolumn{5}{c}{DAQUAR-ALL visual dataset}\\
\hline
Accuracy & 27.54 & 28.43 & 28.62 & 28.79\\
\hline
\end{tabular}
}
\caption{
Test error rates of various model architectures on the \babi-10k dataset, and accuracy performance on the DAQUAR-ALL visual dataset.
The skipped \babi questions (1,4,11,12,13,15,19) achieved 0 error across all models.
}
\label{table:babi-compare}
\end{table}

\subsection{Comparison to state of the art using \babi-10k} \label{sec:babi-sota}

We trained our models using the Adam optimizer \cite{kingma2014adam} with a learning rate of 0.001 and batch size of 128.
Training runs for up to 256 epochs with early stopping if the validation loss had not improved within the last 20 epochs.
The model from the epoch with the lowest validation loss was then selected.
Xavier initialization was used for all weights except for the word embeddings, which used random uniform initialization with range $[-\sqrt{3}, \sqrt{3}]$.
Both the embedding and hidden dimensions were of size $d = 80$.
We used $\ell_2$ regularization on all weights except bias and used dropout on the initial sentence encodings and the answer module, keeping the input with probability $p=0.9$.
The last 10\% of the training data on each task was chosen as the validation set.
For all tasks, three passes were used for the episodic memory module, allowing direct comparison to other state of the art methods.
Finally, we limited the input to the last 70 sentences for all tasks except QA3 for which we limited input to the last 130 sentences, similar to \citet{Sukhbaatar2015}.

On some tasks, the accuracy was not stable across multiple runs.
This was particularly problematic on QA3, QA17, and QA18.
To solve this, we repeated training 10 times using random initializations and evaluated the model that achieved the lowest validation set loss.

\textbf{Text QA Results}

We compare our best performing approach, DMN+, to two state of the art question answering architectures: the end to end memory network (E2E) \cite{Sukhbaatar2015} and the neural reasoner framework (NR) \cite{Peng2015}.
Neither approach use supporting facts for training.

The end-to-end memory network is a form of memory network \cite{Weston2015} tested on both textual question answering and language modeling.
The model features both explicit memory and a recurrent attention mechanism.
We select the model from the paper that achieves the lowest mean error over the \babi-10k dataset.
This model utilizes positional encoding for input, RNN-style tied weights for the episode module, and a ReLU non-linearity for the memory update component.

The neural reasoner framework is an end-to-end trainable model which features a deep architecture for logical reasoning and an interaction-pooling mechanism for allowing interaction over
multiple facts.
While the neural reasoner framework was only tested on QA17 and QA19, these were two of the most challenging question types at the time.

\begin{table}
\centering
\begin{tabular}{lrrr}
Task & DMN+ & E2E & NR \\
\hline
2: 2 supporting facts & 0.3 & 0.3& -\\
3: 3 supporting facts & 1.1 & 2.1& -\\
5: 3 argument relations & 0.5 & 0.8& -\\
6: yes/no questions & 0.0 & 0.1& -\\
7: counting & 2.4 & 2.0& -\\
8: lists/sets & 0.0 & 0.9& -\\
9: simple negation & 0.0 & 0.3& -\\
11: basic coreference & 0.0 & 0.1& -\\
14: time reasoning & 0.2 & 0.1& -\\
16: basic induction & 45.3 & 51.8& -\\
17: positional reasoning & 4.2 & 18.6 & 0.9\\
18: size reasoning & 2.1 & 5.3 & -\\
19: path finding & 0.0 & 2.3 & 1.6\\
\hline
Mean error (\%) & 2.8 & 4.2 & - \\
Failed tasks (err \textgreater 5\%) & 1 & 3 & - \\
\end{tabular}
\caption{
Test error rates of various model architectures on tasks from the the \babi English 10k dataset.
E2E = End-To-End Memory Network results from \citet{Sukhbaatar2015}.
NR = Neural Reasoner with original auxiliary task from \citet{Peng2015}.
DMN+ and E2E achieve an error of 0 on \babi question sets (1,4,10,12,13,15,20).
}
\label{table:babi-sota}
\end{table}

In Table \ref{table:babi-sota} we compare the accuracy of these question answering architectures, both as mean error and error on individual tasks.
The DMN+ model reduces mean error by 1.4\% compared to the the end-to-end memory network, achieving a new state of the art for the \babi-10k dataset.

One notable deficiency in our model is that of QA16: Basic Induction.
In \citet{Sukhbaatar2015}, an untied model using only summation for memory updates was able to achieve a near perfect error rate of $0.4$.
When the memory update was replaced with a linear layer with ReLU activation, the end-to-end memory network's overall mean error decreased but the error for QA16 rose sharply.
Our model experiences the same difficulties, suggesting that the more complex memory update component may prevent convergence on certain simpler tasks.

The neural reasoner model outperforms both the DMN and end-to-end memory network on QA17: Positional Reasoning.
This is likely as the positional reasoning task only involves minimal supervision - two sentences for input, yes/no answers for supervision, and only 5,812 unique examples after removing duplicates from the initial 10,000 training examples.
\citet{Peng2015} add an auxiliary task of reconstructing both the original sentences and question from their representations.
This auxiliary task likely improves performance by preventing overfitting.

\begin{table}[t!]

\centering
\setlength{\tabcolsep}{.5em}
\begin{tabular}{lllrrr}
\hline
\multicolumn{1}{c}{}&
\multicolumn{4}{c}{test-dev}&
\multicolumn{1}{r}{test-std}\\
\cline{2-5}
Method&All&Y/N&Other&Num&All\\
\hline
VQA& & & & &\\
Image&28.1&64.0 &3.8 &0.4 &-\\
Question &48.1&75.7 &27.1 &36.7&-\\
Q+I&52.6&75.6 &37.4 &33.7 &-\\
LSTM Q+I&53.7&78.9 &36.4 &35.2 &54.1\\
\hline
ACK&55.7&79.2 &40.1 &36.1 &56.0\\
iBOWIMG&55.7&76.5 &42.6 &35.0 &55.9\\
DPPnet&57.2&80.7 &41.7 &37.2 &57.4\\
D-NMN&57.9&80.5 &43.1 &37.4 &58.0\\
SAN&58.7&79.3 &46.1 &36.6 &58.9\\
\hline
DMN+&\textbf{60.3}&80.5 &48.3 &36.8 &\textbf{60.4}\\
\end{tabular}
\caption{Performance of various architectures and approaches on VQA test-dev and test-standard data. VQA numbers are from ~\citet{Antol2015};  ACK ~\citet{wu2015ask}; iBOWIMG -\citet{Zhou2015}; DPPnet - \citet{noh2015image}; D-NMN - \citet{andreas2016learning}; SAN -\citet{yang2015stacked} }
\label{quantitative}
\end{table}

\subsection{Comparison to state of the art using VQA} \label{sec:vqa-sota}
For the VQA dataset, each question is answered by multiple people and the answers may not be the same, the generated answers are evaluated using human consensus.
For each predicted answer $a_i$ for the $i_{th}$ question with target answer set $T^{i}$, the accuracy of VQA:
$Acc_{VQA} = \frac{1}{N}\sum_{i=1}^Nmin(\frac{\sum_{t\in T^i}\mathds{1}_{(a_i==t)}}{3},1)$
where $\mathds{1}_{(\cdot)}$ is the indicator function. Simply put, the answer $a_i$ is only 100$\%$ accurate if at least 3 people provide that exact answer.

\textbf{Training Details}
We use the Adam optimizer \cite{kingma2014adam} with a learning rate of 0.003 and batch size of 100.
Training runs for up to 256 epochs with early stopping if the validation loss has not improved in the last 10 epochs.
For weight initialization, we sampled from a random uniform distribution with range $[-0.08, 0.08]$.
Both the word embedding and hidden layers were vectors of size $d=512$.
We apply dropout on the initial image output from the VGG convolutional neural network \cite{simonyan2014very} as well as the input to the answer module, keeping input with probability $p=0.5$.

\begin{figure*}[t!]
   \centering
   	 	\includegraphics[width=0.95\textwidth]{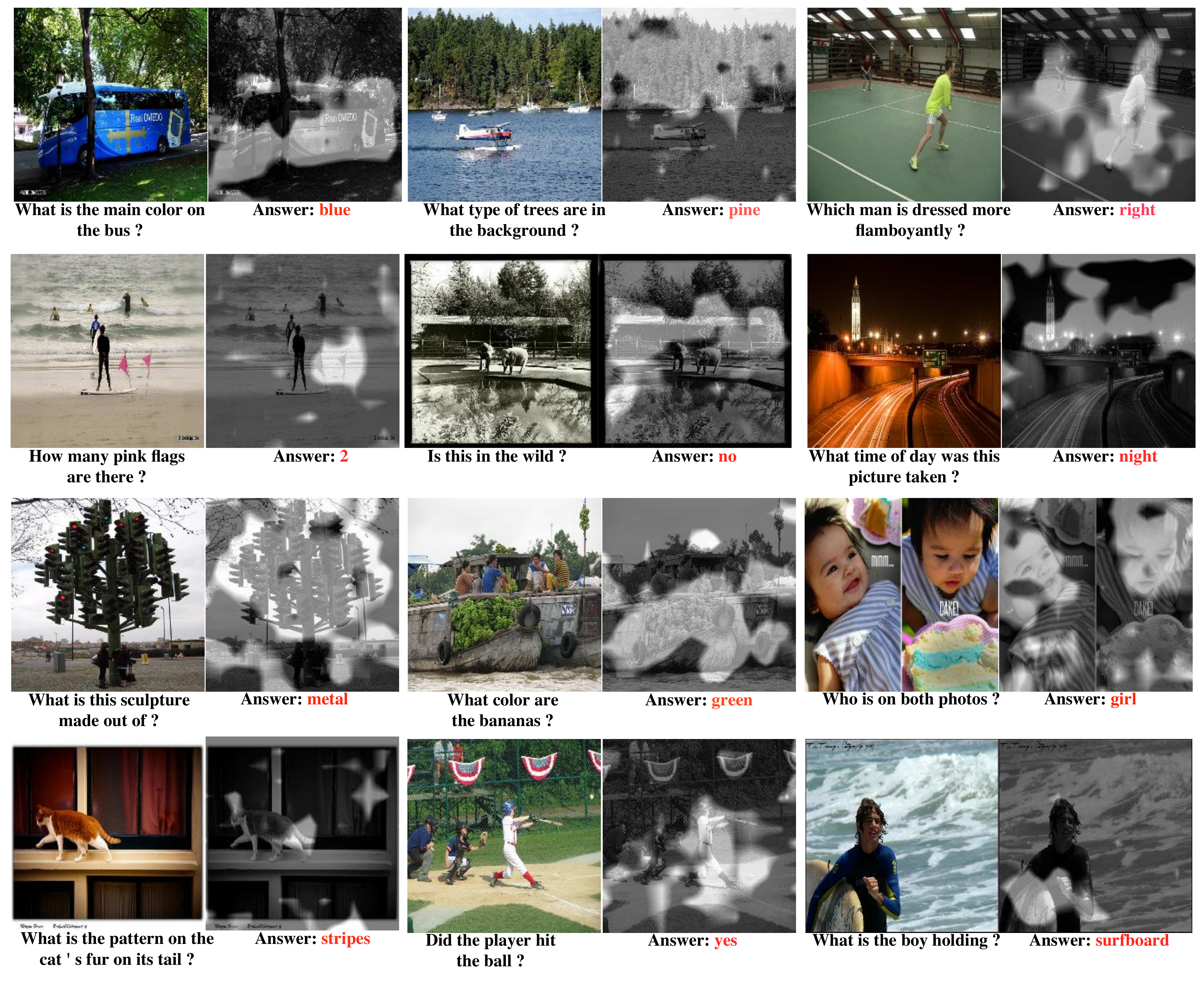}
   	 	\vspace{-0.6cm}
   \caption{
   Examples of qualitative results of attention for VQA.
   The original images are shown on the left.
   On the right we show how the attention gate $g^t_i$ activates given one pass over the image and query.
   White regions are the most active.
   Answers are given by the DMN+. }
   	 	\vspace{-0.2cm}
   \label{fig:qualitative}
\end{figure*}

\textbf{Results and Analysis}

The VQA dataset is composed of three question domains: Yes/No, Number, and Other.
This enables us to analyze the performance of the models on various tasks that require different reasoning abilities.

The comparison models are separated into two broad classes: those that utilize a full connected image feature for classification and those that perform reasoning over multiple small image patches.
Only the SAN and DMN approach use small image patches, while the rest use the fully-connected whole image feature approach.

Here, we show the quantitative and qualitative results in Table~\ref{quantitative} and Fig.~\ref{fig:qualitative}, respectively.
The images in Fig.~\ref{fig:qualitative} illustrate how the attention gate $g^t_i$ selectively activates over relevant portions of the image according to the query.
In Table~\ref{quantitative}, our method outperforms baseline and other state-of-the-art methods across all question domains (\textbf{All}) in both test-dev and test-std, and especially for \textbf{Other} questions,  achieves a wide margin compared to the other architectures, which is likely as the small image patches allow for finely detailed reasoning over the image.

However, the granularity offered by small image patches does not always offer an advantage. The Number questions may be not solvable  for both the SAN and DMN architectures, potentially as counting objects is not a simple task when an object crosses image patch boundaries.


\section{Conclusion}
We have proposed new modules for the DMN framework to achieve strong results without supervision of supporting facts.
These improvements include the input fusion layer to allow interactions between input facts and a novel attention based GRU that allows for logical reasoning over ordered inputs.
Our resulting model obtains state of the art results on both the VQA dataset and the \babi-10k text question-answering dataset, proving the framework can be generalized across input domains.


\bibliography{allBibsFinal}
\bibliographystyle{icml2016}

\end{document}